%% file: main.tex
\documentclass[conference]{IEEEtran}
\usepackage{blindtext, graphicx}
\usepackage{listings}
\usepackage{graphicx}
\usepackage[font=small]{caption}

\usepackage[utf8]{inputenc}
\usepackage{tikz}
\usepackage{float}

\newcommand{\oautor}{{Source: the Authors, 2024.}}
\def\checkmark{\tikz\fill[scale=0.3](0,.35) -- (.25,0) -- (1,.7) -- (.25,.15) -- cycle;}
\usepackage{relsize,etoolbox}
\AtBeginEnvironment{quote}{\smaller}

\newfloat{floatquote}{tbp}{loq}
\floatname{floatquote}{Snippet}
\usepackage{caption}
\usepackage{cleveref}
\crefname{floatquote}{snippet}{snippets}
\Crefname{floatquote}{Snippet}{Snippets}

\title{Chattronics: using GPTs to assist in the design of data acquisition systems}

\author{
\IEEEauthorblockN{Jonathan Paul Driemeyer Brown}
\IEEEauthorblockN{Tiago Oliveira Weber}

\IEEEauthorblockN{}
\IEEEauthorblockN{Federal University of Rio Grande do Sul}
}

\begin{document}

\maketitle
\thispagestyle{empty}
\pagestyle{empty}

\begin{abstract}

The usefulness of Large Language Models (LLM) is being continuously tested in various fields. However, their intrinsic linguistic characteristic is still one of the limiting factors when applying these models to exact sciences. In this article, a novel approach to use General Pre-Trained Transformers to assist in the design phase of data acquisition systems will be presented. The solution is packaged in the form of an application that retains the conversational aspects of LLMs, in such a manner that the user must provide details on the desired project in order for the model to draft both a system-level architectural diagram and the block-level specifications, following a Top-Down methodology based on restrictions.  
To test this tool, two distinct user emulations were used, one of which uses an additional GPT model. In total, 4 different data acquisition projects were used in the testing phase, each with its own measurement requirements: angular position, temperature, acceleration and a fourth project with both pressure and superficial temperature measurements.
After 160 test iterations, the study concludes that there is potential for these models to serve adequately as synthesis/assistant tools for
data acquisition systems, but there are still technological limitations. The results show coherent architectures and topologies, but that GPTs have difficulties in simultaneously considering all requirements and many times commits theoretical mistakes.


\end{abstract}

\section{Introduction}

Large Language Models (LLM) have proven to be capable of achieving impressive milestones, such as passing the bar exam  \cite{law}, medical diagnosis \cite{sallam2023chatgpt} and code optimisation \cite{biswas2023role}. 
Applications such as these show that the adoption of LLMs has been faster in fields in which text-based outputs can be directly used. In fields such as Electrical and Electronic Engineering, additional steps are required to properly leverage natural language and allow LLMs to reliably help.

The application is packaged as a command line interface (CLI) and follows a pre-established conversational structure, which requests the user to answer questions and interact with the model. The goal of such flow is to limit the scope and help the model understand better the project requirements, in order to improve the final proposed solution. 

One of the main challenges of basing applications on LLMs is automating the testing process.
The non-deterministic characteristic of these models makes it difficult to run multiple iterations of the same scenario and expect the same output. 
Given that the tool requires user interaction - as the model will ask questions and request feedback - automated tests have to be able to sufficiently emulate human behaviour. For that, two approaches were developed, one of which uses a second GPT model with its own set of prompts. 

For each user emulation method and testbench - which consists of projects based on data acquisition \cite{Rosemary1997} and instrumentation textbooks \cite{balbinot} - 20 iterations of the tests were run, totalling 160 iterations. All iterations were manually analysed, in order to identify the different types of errors and how many times they appear across iterations.

Currently, system and block-level design of data acquisition systems is mostly manual. Although some blocks, such as filters, can be designed using specific tools, high-level decisions and the specification translation from system-level to block-level is performed by experienced designers. Both for educational and industrial purposes, the use of automated strategies can benefit parts of the design flow. In education, such tools can help students see how system-level specifications are converted to block-levels and understand the rationale of the design flow. For the industry, this type of tool can be used in non-demanding projects, saving time for experienced designers and improving the time-to-market of products.

LLMs were already used to assist the synthesis of digital chips \cite{ChipChat}, in which a chip design was co-authored by a transformer model, in this case, ChatGPT-4.

ChatGPT has also been used to assist in the field of embedded systems \cite{ChatGPTElectronics}, using the model to provide specifications on the controller, sensor models, synthesis flow and helpful literature.
The conclusion is that much of the answers still require validation by an expert given eventual inconsistencies.

These papers manually use and manipulate GPT in order to get their answers. In order to develop a tool that can be packaged and distributed, however, more prompt engineering and conversational flow techniques have to be incorporated into the application. These are very common in open-source projects \cite{gpt-pilot}\cite{Osika_gpt-engineer_2023}, which use GPTs conversational back-and-forth capabilities to build a solution piece by piece. 

This paper incorporates frequently used techniques in LLM based tools, such as personas, requesting questions from the model and templates, in order to develop an application for Data Acquisition System project assistance. Specific testing techniques, which propose methods to emulate the user, and testbenches are also presented. The results were manually analysed and the main takeaways will be discussed.

\subsection{Contributions}
\begin{itemize}
    \item As far as the authors are aware, this is the first work to apply Large Language Models in the field of Data Acquisition Systems;
    \item the work proposes a novel approach to automated testing using the interaction of two GPT models (the designer and the user emulator);
    \item four testbenches -  based on textbooks in the field \cite{Rosemary1997}\cite{balbinot}, each composed of a project description and a list of requirements, that allow the testing of the model's capabilities in qualitative, quantitative and diagramming aspects;
    \item  the development of an automated tool for designing data acquisition system using LLMs, receiving high-level specifications and generating the architecture, block-level specifications and generating a system diagram.
\end{itemize}

\section{Methodology}
\subsection{Materials}
The application was developed in Go using the \textit{go-openai} library, which acts as an SDK (Software Development Kit) to OpenAI official API (Application Programming Interface). The model used for both GPTs in this paper was OpenAI's GPT-4-Turbo \cite{gpt4turboAnnouncement}, officially named \textit{gpt-4-1106-preview}, as it was in preview version at the time. The supposed better performance matched with the lower cost of this model made it a viable option, despite not being an official major version of GPT. The only configured parameter was temperature, which was set to 0.8 for the main GPT and 0.2 for the user emulating model.

\subsection{Chat Flow}
The conversational flow, i.e. the planned back and forth between the model and the client (be it a user or an emulator), aims to reduce superficial answers by the model that might not satisfy all project requirements. As the application was developed with the purpose of being an interactive tool, it must consider the case in which the user does not initially provide enough information to form a satisfactory solution. 

\begin{figure}[H]
    \caption{Diagram with the chat flow overview.}
    \centering
    \includegraphics[width=\columnwidth]{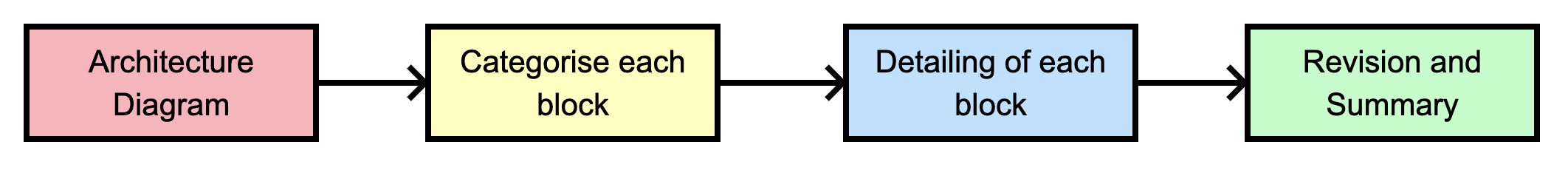}
    \label{fig:overviewChatDiagram}
\oautor
\end{figure}

A high level diagram showing the stages included in the application's chat flow can be seen in Figure \ref{fig:overviewChatDiagram}.
Given that the chosen approach was system level synthesis with top-down methodology, the first step is to establish the architecture of the proposed solution. Following it, each block should be individually analysed in order to establish lower level requirements and details. The stages named \textit{Categorisation} and \textit{Revision} are mostly for internal purposes and will be explained in the next sections.

\subsubsection{Architectural Stage}
\begin{figure}[H]
    \caption{Architectural stage chat flow.}
    \centering
    \includegraphics[width=\columnwidth]{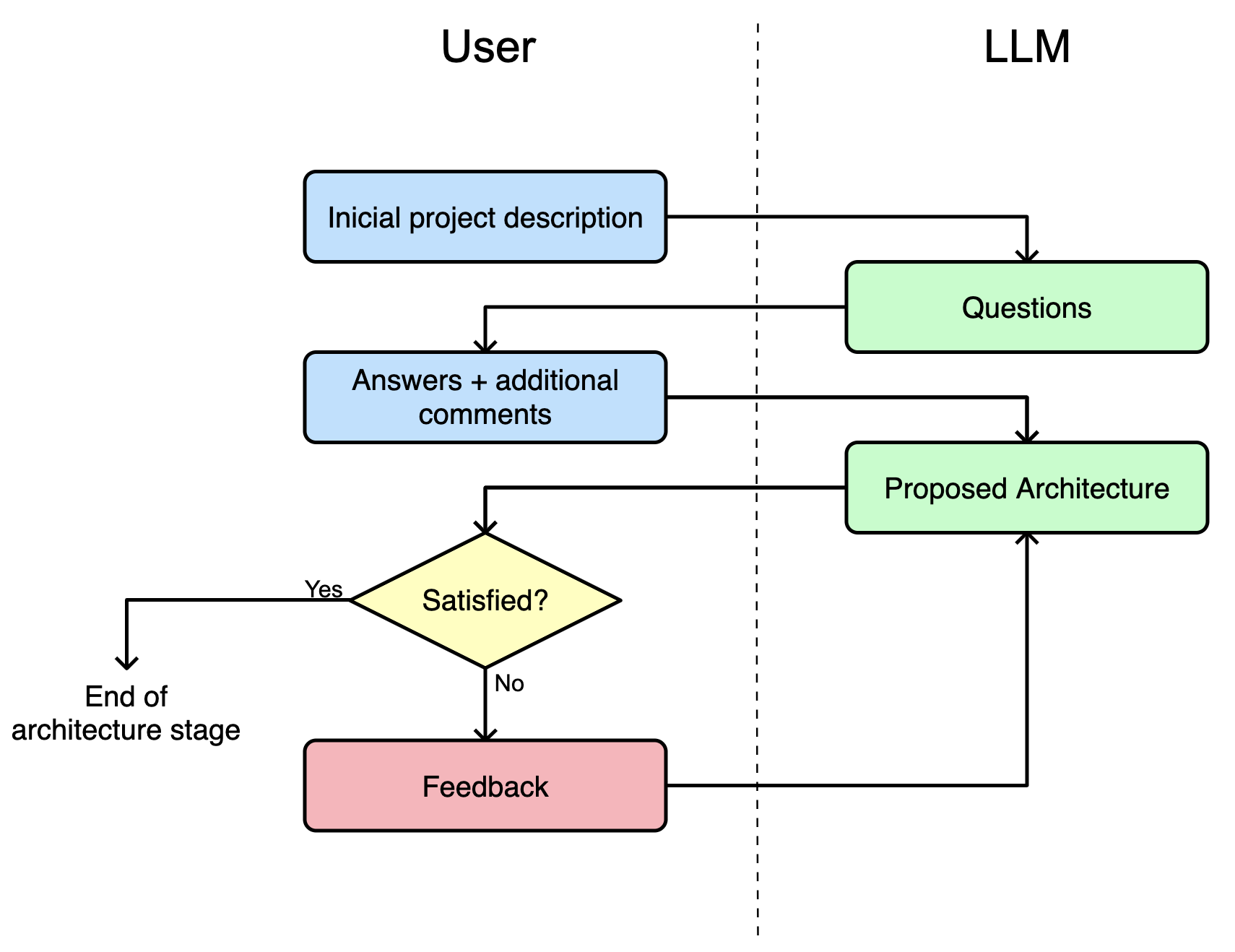}
    \label{fig:architectureChatDiagram}
\oautor
\end{figure}

This stage is the application's entry point and aims to gather the main system requirements of the project in order to design a block diagram representing the architecture. Figure \ref{fig:architectureChatDiagram} shows the internal steps of this stage, which starts with the user input that describes the desired project. The next step is an exchange of questions and answers, in which the model will ask up to 5 questions to further understand the provided project description. After the user (or emulator) provides the answers - or leaves the answers empty, given that in some cases the user may not have the requested information -, the model will generate a .DOT string of the architecture, which the application visually presents to the user. 

If the user is not satisfied, a loop begins in which the model takes the user's feedback to generate a new version of the diagram until it's acceptable and the stage ends. As GPT does not have memory, it's necessary to resend the entire conversation in order to emulate it. However, the application does not always include an exchange in future prompts, as it might only increase the already large size of the conversation and confuse the model. After the feedback loop, only the final and accepted architecture is included in the conversation history, excluding all failed attempts.

\subsubsection{Categorisation Stage}
This stage is entirely internal, i.e. without interaction with the client, and the main purpose is to set up the application for the detailing phase. Given that each individual block has to go through its own detailing step, the prompts would have to be very generic given that the block in question could be either a sensor, an analogue-digital converter (ADC) or a low-pass filter, etc. 

Hence the categorisation stage, that attributes a category to each block of the designed architecture. This enables each category to have it's own set of prompts, allowing the ADC stage to specifically request the model to provide a sampling rate, for example. Each block can be assigned to one of the following categories:

\begin{itemize}
    \item Sensor;
    \item Signal conditioning;
    \item Amplification;
    \item Filtering;
    \item Other conditioning;
    \item Direct measurement;
    \item Analogue-digital converter;
    \item Digital processing;
    \item Others.
\end{itemize}

The categories named \textit{others} and \textit{other conditioning} serve as fallback in case none of the other ones fit.
The main target of the project is on the analogue side and not the digital part, which explains the more generic \textit{digital processing} category in comparison to the rest.

\subsubsection{Detailing Stage}
The detailing stage follows a similar flow as to the architectural one, with the only difference being that the input text that will prompt the model to ask the questions is not user generated, but rather triggered by the application of a specific prompt for each category. Other than that, the question and answer steps still remain, as well as the feedback loop, which depends on the users satisfaction with the provided detailing of the block. 

In some cases, the amount of blocks in the architectural diagram can be large, which makes the conversation history scale up when adding up all interactions in the detail phase. This is not only a problem cost wise, but it might also affect the attention given to specific parts of the input, lowering the value of otherwise important information. This resulted in the decision not to share information across detail stages, i.e. the conversation history of every detail stage is only the architectural conversation and does not contain the detailing of previous blocks.  
This proved to significantly shape the outcome of the solution, as will be discussed in later sections of the paper.

\subsubsection{Revision}
The final stage is where the user will be able to see the proposed solution as a whole and request corrections if necessary. During preliminary tests, it was observed that each block generated almost a pages worth of detailing, causing the final solution to be excessively large. Despite the attempts to mitigate GPT's prolix writing, a simple concatenation of all generated text would be too long and impractical. Therefore, after all blocks are finished being detailed, the model compiles everything into a summary, but is instructed to keep all numerical values, equations, parameters and variables that were generated, to reduce the chance of losing important information. 

The user is presented with this summary and is asked whether something must be altered, following the same feedback loop method used in the previous stages. This is the last chance to request changes to the solution, as it marks the end of the process as a whole.

\subsection{Automated Testing}
To run automated tests on the application it was necessary to find ways to emulate the user. Given that there is no way of knowing what questions the model would ask and what phrasing will be used, hardcoding the answers was not an option. An exception to this is if the emulated user actually does not provide any answers, making the initial project description the only information provided to the model. This will be called the \textit{Direct Context} approach.

The other alternative is to instruct another GPT model to emulate the user, providing it with a set of information that it can answer from if asked. If the list says that the output voltage must be between 0 and 20 Volts, for example, the user-emulating model can provide this as an answer if - and only if - that's the questions asked by the application. This will be called the \textit{Open Context} approach and it not only tests the application's ability to generate adequate solutions, but also its ability to extract the required restrictions to build such solutions.

To allow a fair comparison of both approaches, the list provided to the \textit{open context} is included in the project description used by the \textit{direct context}, as if all project requirements are passed straight away to the model and excusing it of the need to ask the questions. Table \ref{tab:comparacao_contextos} compares both approaches.

\input{compare_contexts}

\subsection{Testbenches}
To test different capabilities of the application, a set of 4 testbenches were chosen, each of which requires a specific expertise to solve. They were based on data acquisition projects \cite{Rosemary1997} and instrumentation exercises \cite{balbinot}. Each testbench is composed of 2 prompts: a project description and a list of possible answers. With them it's possible to build all required testing prompts. 

\subsubsection{Angular Position Project}
This testbench \cite{balbinot} does not require extensive calculations, nor does it tackle more advanced forms of sensoring: it's a potentiometer, whose output voltage must be measured by a Data Acquisition device. The main challenge of this project is to satisfy two important requirements, the maximum input voltage on the DAQ must not exceed the recommended limit and frequencies near the 50 and 60 Hz must be dampened. The project description, as well as the list of requirements, can be seen in
\cref{desc_pot}.

\begin{quote}
I want to develop a project that calculates the angle of a pendulum by using a potentiometer.
The component will be fixed to a wooden structure and another beam will be connected to the adjustable pin.
The analog output will be sent to a DAQ which cannot be altered or tuned.
Use the simplest design possible, which takes into account all requirements.\\
1. The potentiometer is linear;\\
2. The value of the potentiometer is 10 kOhms;\\
3. The power source connected to the potentiometer is between -10 and 10 Volts;\\
4. The angle range that the potentiometer should be able to measure is from 45 to 135 degrees, in which 90 degrees corresponds to the steady position;\\
5. The sampling rate of the DAQ is 1000 samples per second;\\
6. Frequencies around 50 and 60 Hz should be attenuated;\\
7. The conditioning should be done in the analog domain;\\
8. The DAQ is only for acquisition, you can not alter its configuration;\\
9. The maximum accepted input voltage for the DAQ is +/- 7 V.
\captionof{floatquote}{Project description and requirements list for the angular position project.}
\label{desc_pot}
\end{quote}

This project serves to evaluate whether the application can handle simpler requests, but also if it's capable of providing numerical values to dampen and limit the signal. A possible architecture for such project consists of the sensoring block, followed by amplification and filtering, ending with the DAQ.

\subsubsection{Thermometry Project}
This testbench \cite{balbinot} focuses on the qualitative aspects of the projects, without requesting specific numeric values. Unlike the previous testbench, this one requires more conditioning blocks in order to enable a viable implementation of the solution. Given that the sensor is an NTC, which is non-linear, a linearization step is mandatory as well as a Wheatstone bridge. There are alternatives to this, but complicated solutions also impact viability. The project description, as well as the list of requirements, can be seen in
\cref{desc_term}.

\begin{quote}
I want to develop a project that measures the temperature of water inside a beaker with a thermistor and that outputs a voltage that can be easily measured by a multimeter. \\
1. The output voltage range to be measured by the multimeter should be between 0 and 20 Volts;\\
2. The temperature that will be measured is between 10 and 90 degrees celsius;\\
3. The self heating effect must be considered and must be less than 1 percent;\\
4. There should be no ADC other than the one inside the multimeter (which can't be changed);\\
5. All conditioning should be done in the analog domain;\\
6. The sensor to be used is an NTC Vishay NTCLE100E3;\\
7. The NTC must be linearized for the midpoint of the input range by a resistor in parallel.
\captionof{floatquote}{Project description and requirements list for the thermometry project.}
\label{desc_term}
\end{quote}

\subsubsection{Portable Accelerometry for Low-Frequency Vibrations}
This project \cite{Rosemary1997} differs from the other testbenches as it focuses on the calculations and numeric values associated to a project. In the book, the author recognises that the use of a piezoelectric sensor might not be ideal for such applications, but provides the step by step guide to build such a solution. 

The information inside the answers list - which is directly provided in the \textit{direct context} approach - indicates the numeric values that the solution should satisfy. Therefore, the requirements list validates whether the component, parameters and project choices lead to such values. The architecture can contain only a charge amplification stage between the sensor and the signal acquisition, or use an additional voltage gain block. An important note to this testbench is that the signal acquisition stage is mostly overlooked, and the application is instructed by the description to focus on the conditioning. The project description, as well as the list of requirements, can be seen in
\cref{desc_acc}.

\begin{quote}
Design a portable device capable of measuring a low-frequency vibration.
Make sure to calculate all components and overall values involved in the project, including maximum bias current, charge amplifier components considering a peak-to-peak voltage output.\\
1. Use a piezoelectric accelerometer;\\
2. The accelerometer has a sensitivity of 100 pC/g (g is the acceleration due to gravity);\\
3. Use the peak to peak acceleration to calculate charge;\\
4. An input oscillation has a frequency up to 2 Hz;\\
5. The input oscillation has a peak amplification of 5 cm;\\
6. Output voltage should be 1 V peak to peak;\\
7. Low frequency response of the system has to be 3 dB down at 0.25 Hz;\\
8. The offset should be kept as less than 10 mV.
\captionof{floatquote}{Project description and requirements list for the accelerometry project.}
\label{desc_acc}
\end{quote}

\subsubsection{Machinery Pressure and Temperature Monitor}
The only testbench to include more than one sensing principle, this project \cite{Rosemary1997} is unique as it contains multiple channels and requires the application to provide specific conditioning for the strain gauges and for the non-linear temperature sensors. This testbench contains the most project requirements, a consequence of the larger block diagram that must be developed for it. The project description, as well as the list of requirements, can be seen in \cref{desc_machine}.

\begin{quote}
Design a system to monitor pressure and surface temperature variations at eight points of a machine and the data is to be analysed by computer.
Design the acquisition system but don't worry about any software implementations or logic analysis.\\
1. The accuracy of both sensors should be at 1 percent;\\
2. The pressure sensors have a maximum output of 1 mV;\\
3. Temperature sensors are radiation detectors having a maximum output of 100 mV;\\
4. Temperature sensors have a nonlinear scale;\\
5. Strain-gauge should be used as the pressure sensor;\\
6. Infrared radiation detectors should be used for the surface temperature sensor;\\
7. Assume all pressure sensors have identical sensitivity;\\
8. Signal frequencies up to 400 Hz are expected;\\
9. Frequencies higher than 400 Hz (if present) need not be recorded;\\
10. The ADC should sample one channel at a time.
\captionof{floatquote}{Project description and requirements list for the pressure and superficial temperature project.}
\label{desc_machine}
\end{quote}

The main goal of this testbench is to evaluate whether the model can provide a suitable architecture to a project that involves many steps. Not only must the diagram cover the non-linear conditioning of 8 parallel strain gauges, which require Wheatstone bridges and instrumentation amplifiers, but also conditioning of 8 parallel temperature sensors, that need linearization. All of which need to be subject to an anti-aliasing filter before being sampled by an ADC. 

\begin{figure}[H]
    \caption{Example architecture provided by \textit{Data Acquisition Systems}.}
    \centering
    \includegraphics[width=\columnwidth]{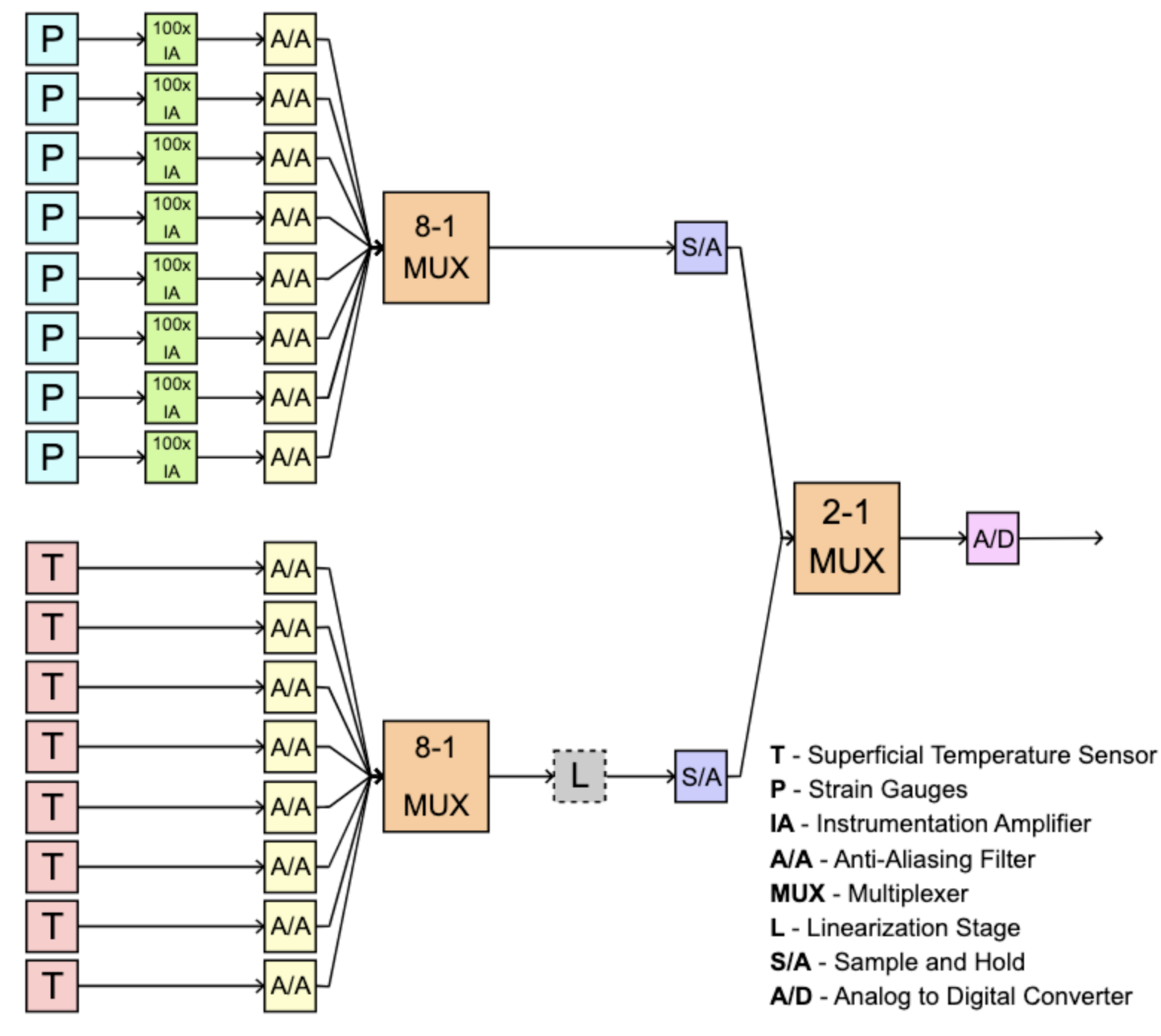}
    \label{fig:testbench_arq_rosemary}
{Source: Adapted from Taylor (1997) \cite{Rosemary1997}}
\end{figure}

The proposed architecture can be seen in Figure \ref{fig:testbench_arq_rosemary}.

\section{Results}

\subsection{Best Solutions per Testbench}
Each project testbench was synthesised 20 times in the direct context mode (without question and answer interactions) and 20 times in the \textit{open} context mode (with another model to emulate the user interaction), resulting in a total of 40 synthesis per testbench and 160 considering all testbenches.

\begin{figure}[H]
    \caption{Diagram of the best solution for the angular position testbench.}
    \centering
    \includegraphics[width=\columnwidth]{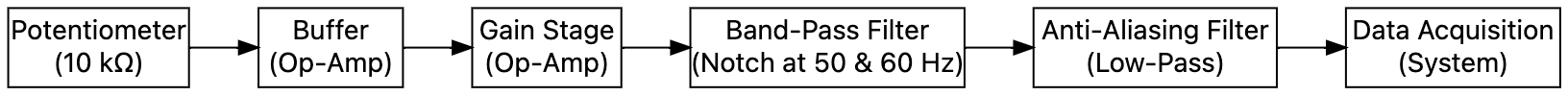}
    \label{fig:best_pot}
\oautor
\end{figure}

For the angular position testbench, the best solution can be found in the seventh iteration with \textit{direct context} and the diagram can be seen in Figure \ref{fig:best_pot}. The textual solution provided by the model - adapted for better visualisation - can be seen in \cref{res_pot}. In it, the model correctly limits the maximum voltage, not only by correctly adjusting the voltage gain, but also by using safety diodes, which is a plus. Instead of using a low-pass filter, which would filter out 50 and 60 Hz frequencies, the LLM opted for two Notch filters, forming a rejection band - which it incorrectly named a pass band.
Other than the naming mistake, there were no theoretical errors. The use of a buffer can be considered unnecessary, however.

\begin{quote}
Potentiometer (Sensor)\\
- Type: Linear rotary potentiometer.\\
- Value: 10 kOhms.\\
- Model Suggestion: Bourns 3590 Precision Potentiometer.\\
- Mounting: Securely attached to the wooden structure with the shaft coupled to the pendulum beam.\\
- Voltage Range: Expected to vary linearly with the angle of the pendulum, estimated between -5V and +5V for 45° to 135°.\\

Buffer (Signal Conditioning)\\
- Topology: Voltage Follower Configuration using an Operational Amplifier (Op-Amp).\\
- Op-Amp Suggestion: Texas Instruments OPA340 or Analog Devices AD8605.\\
- Power Supply: ±10V (matching the potentiometer's power supply).\\
- Input and Output Resistances: High input impedance (>1 M$\Omega$) and low output impedance (<100 $\Omega$) to prevent loading and ensure signal integrity.\\

Gain Stage (Signal Conditioning)\\
- Topology: Non-Inverting Op-Amp Configuration.\\
- Required Gain (A): $( A = \frac{V_{out,max} - V_{out,min}}{V_{in,max} - V_{in,min}} = \frac{7V - (-7V)}{5V - (-5V)} = 1.4 )$.\\
- Resistor Values: $R1 = 4k\Omega$, $R2 = 10k\Omega$ (precision resistors with a 1\% tolerance).\\
- Power Supply: ±10V (dual supply).\\

Band-Pass Filter (Signal Conditioning)\\
- Active Twin-T Notch Filter Topology for 50 Hz and 60 Hz attenuation.\\
- Notch Filter Component Values: $R = 31.8 k\Omega$, $C = 100 nF$ (for 50 Hz), $R = 26.5 k\Omega$, $C = 100 nF$ (for 60 Hz).\\
- Followed by a Sallen-Key Low-Pass Filter with a Butterworth response.\\
- Low-Pass Filter Component Values: $R = 3.9 k\Omega$, $C = 1 nF$ (for a cutoff frequency of 200 Hz).\\

Anti-Aliasing Filter (Signal Conditioning)\\
- Sallen-Key Low-Pass Filter Topology.\\
- Filter Order: Second-order (two-pole) with -12 dB/octave roll-off.\\
- Cutoff Frequency: Approximately 210 Hz (below half the DAQ's sampling rate).\\
- Damping Factor: $\xi = 0.707 $(Butterworth response for flat passband).\\

Data Acquisition System (DAQ)\\
- ADC Topology: Successive Approximation Register (SAR).\\
- ADC Resolution: 12 bits (4096 discrete levels).\\
- SNR: 60-70 dB (corresponding to a dynamic range of about 10-11 bits).\\
- Input Capacitance and Leakage Current: Low to prevent loading effects.\\
- Power Supply Voltage: Typically 3.3 V to 5 V for digital circuitry.\\
- Sampling Rate: 1000 samples per second.\\
- Maximum Input Voltage: ±7 V.\\

General Considerations**\\
- All Op-Amps should have decoupling capacitors to reduce noise.\\
- Protection diodes may be used at the buffer output to protect the DAQ from voltage spikes.\\
- A compact design using SMD components is preferred for modern applications, but through-hole components may be used for prototyping ease.\\
- The entire system must be calibrated and tested to ensure accurate angle measurement within the operating voltage and angle range.

\captionof{floatquote}{Project solution suggestion provided by the model in the seventh iteration of the direct context in the angular position testbench.}
\label{res_pot}
\end{quote}

\begin{figure}[H]
    \caption{Diagram of the best solution for the thermometry testbench.}
    \centering
    \includegraphics[width=\columnwidth]{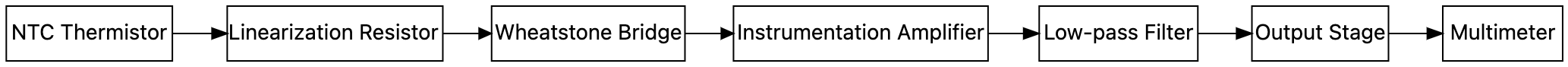}
    \label{fig:best_term}
\oautor
\end{figure}

For the thermometry project, the best solution (see Figure \ref{fig:best_term} of the third iteration with \textit{open context}) contains some inconsistencies, such as adding gain to the \textit{output stage}, when the necessary gain was already correctly set in the \textit{instrumentation amplifier} block. This block inconsistency phenomenon is seen across many iterations for all testbenches and it will be further discussed later on in the paper. As the focus of this project was to test the qualitative aspects of a data acquisition system, it can be considered that the solution satisfied most of the requirements, such as using a Wheatstone bridge, linearizing the NTC, using a differential amplifier to add gain and filtering the signal.

The accelerometry testbench presented the worst solutions, in the sense that few requirements managed to be satisfied and no solution is viable and free of theoretical mistakes. This, however, is not entirely unexpected, given that the goal of this project was to test the calculating capabilities of the model. 

The best iteration was the 18th using the \textit{direct context} approach. 
One of the flaws in the solution was placing the ADC before the anti-aliasing filter, which is enough to make the project unfeasible. 
However, incurring in this error, this synthesis iteration successfully calculated the block-level parameters for the peak-to-peak output voltage to be 1 V.
Even so, this was the only successful calculation as the solution does not correctly satisfy the frequency requirements of the charge amplifier. The inability of the application to provide acceptable solutions to this testbench will be discussed later on.

\begin{figure}[H]
    \caption{Diagram of the best solution for the pressure and temperature monitor testbench.}
    \centering
    \includegraphics[width=\columnwidth]{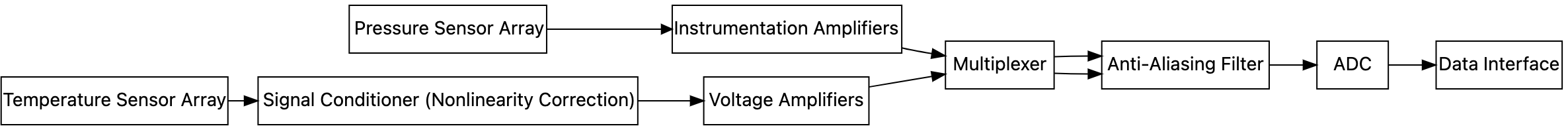}
    \label{fig:best_machine}
\oautor
\end{figure}

Finally, for the pressure and temperature monitor testbench, the diagram of the best solution can be seen in Figure \ref{fig:best_machine}, which refers to the 8th iteration with \textit{direct context}. It successfully provided the gain and cutoff frequencies, without committing theoretical errors. The diagram, however, should explicitly add a Wheatstone bridge block after the strain gauges (it was only mentioned in the detailing) and use double-ended arrows to indicate the differential signal between it and the instrumentation amplifiers.

\subsection{Manual Validation}
Despite there being 160 iterations, many mistakes and decisions follow patterns that are caused by similar problems. For all testbenches, especially for the accelerometry project, the lack of sharing the detailing stage across blocks hindered the consistency across the solution. This was observed in the best solution in the thermometry project when the gain was set twice: in the instrumentation amplifier and in the output stage. 
Despite the block diagram (provided by the tool) already specifying the use of an instrumentation amplifier to adjust the gain, the lack of communication across blocks allows for such mistakes.

In the case of the accelerometry project, the decision to not share the chat history does not allow the blocks to share parameters, many of which must be used in multiple stages in the architecture to successfully apply the equations. Even though many mistakes were still observed in the calculations, regardless of this detail, it's undeniable that the blocks require shared information to better formulate the solution.

In all cases, the application is instructed to provide component models (as can be seen in \cref{desc_pot,desc_term,desc_acc,desc_machine}), non-optimal solutions were detected. In 24 of the 40 iterations of the angular position project, the solution proposed the use of \textit{TL072} Operational Amplifiers (OPAMP). These are not usually used for precision applications and are considered outdated, but that's not enough to disregard the solution, as no requirement was set for such. 

On the other hand, the angular position testbench contained theoretical errors regarding the non-inverting amplifier. In 7 of the 40 iterations, the solutions used such topology with the aim of achieving a gain inferior to 1, which is mathematically impossible. In these cases, the step-by-step of the calculations show that the solution describes an inverting topology - which would be acceptable given that the description does not specify polarity -, but calls it non-inverting.

Still referring to this testbench, 14 iterations of the solutions provided the component values to build the Butterworth low-pass filters and in 11 of those the values were correct. In the remaining three, the values result in a cutoff frequency 10 times greater or lower than expected. Preliminary tests also support the evidence that many calculation mistakes are due to the scale.

The angular position project also shows that GPT has difficulties in implicitly satisfying requirements. The testbench explicitly asks for 50 and 60 Hz to be dampened, but instead of using a single low-pass filter, which would solve both requirements, the model prefers to use two Notch filters. In 3 iterations, the model used three filters, the two Notches and an anti-aliasing. This decision is not incorrect, but outlines a characteristic in how the model provides the solution.

When comparing the \textit{direct} and \textit{open} contexts, an error observed in 7 iterations of the \textit{open context} is the value of the cutoff frequency of the anti-aliasing or low-pass filter, such as setting it to 100 Hz to dampen 50 and 60 Hz. This is believed to be the case when the model did not manage to ask the correct questions in order to get the requirement that requested the dampening of those frequencies. Another inconsistency found in 4 iterations was the positioning of the anti-aliasing filter not being directly before the ADC and in another 4, the cutoff frequency of said filter was close to the desired sampling rate, which is incorrect.

\begin{figure}[H]
    \caption{Incorrect diagram found in the tenth iteration with \textit{direct context}.}
    \centering
    \includegraphics[width=\columnwidth]{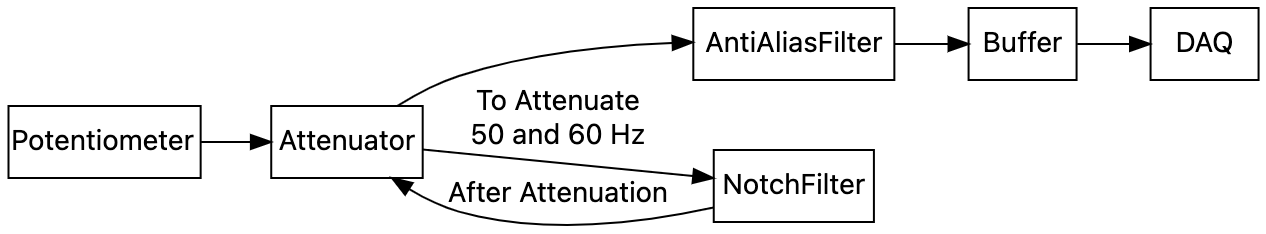}
    \label{fig:results_pot_weird}
\oautor
\end{figure}

In terms of the block diagram for this testbench, most of the iterations provided a sufficiently acceptable architecture, but some, such as in Figure \ref{fig:results_pot_weird}, are incorrect. A mistake found in all testbenches was the incorrect usage of double and single ended arrows, but compared to other mistakes this can be considered less important.

For the thermometry project, in 5 iterations a Wheatstone bridge was paired with a single-ended amplifier topology, and in 4 cases the opposite was done, i.e. a differential topology after a single-ended signal. On the other hand, in 9 iterations the solution describes the correct flow: linearization, bridge and differential gain.

The main issue with the accelerometry testbench was how closely dependent each requirement was with one another. In order for any of the calculations to be successful, the peak-to-peak acceleration should be considered in the first stage of the architecture, and the result should be propagated to the rest of the blocks. As discussed previously, the lack of conversation history hindered this, but when analysing the \textit{open context} approach, if the model was incapable of extracting this requirement from the user, all future restrictions would not be met. This is seen in 17 iterations, in which the model chose an arbitrary value for the acceleration. 
None of the solutions in this testbench managed to satisfy all project requirements.

\begin{figure}[H]
    \caption{Diagram without the usage of arrays, found in the preliminary tests.}
    \centering
    \includegraphics[width=\columnwidth]{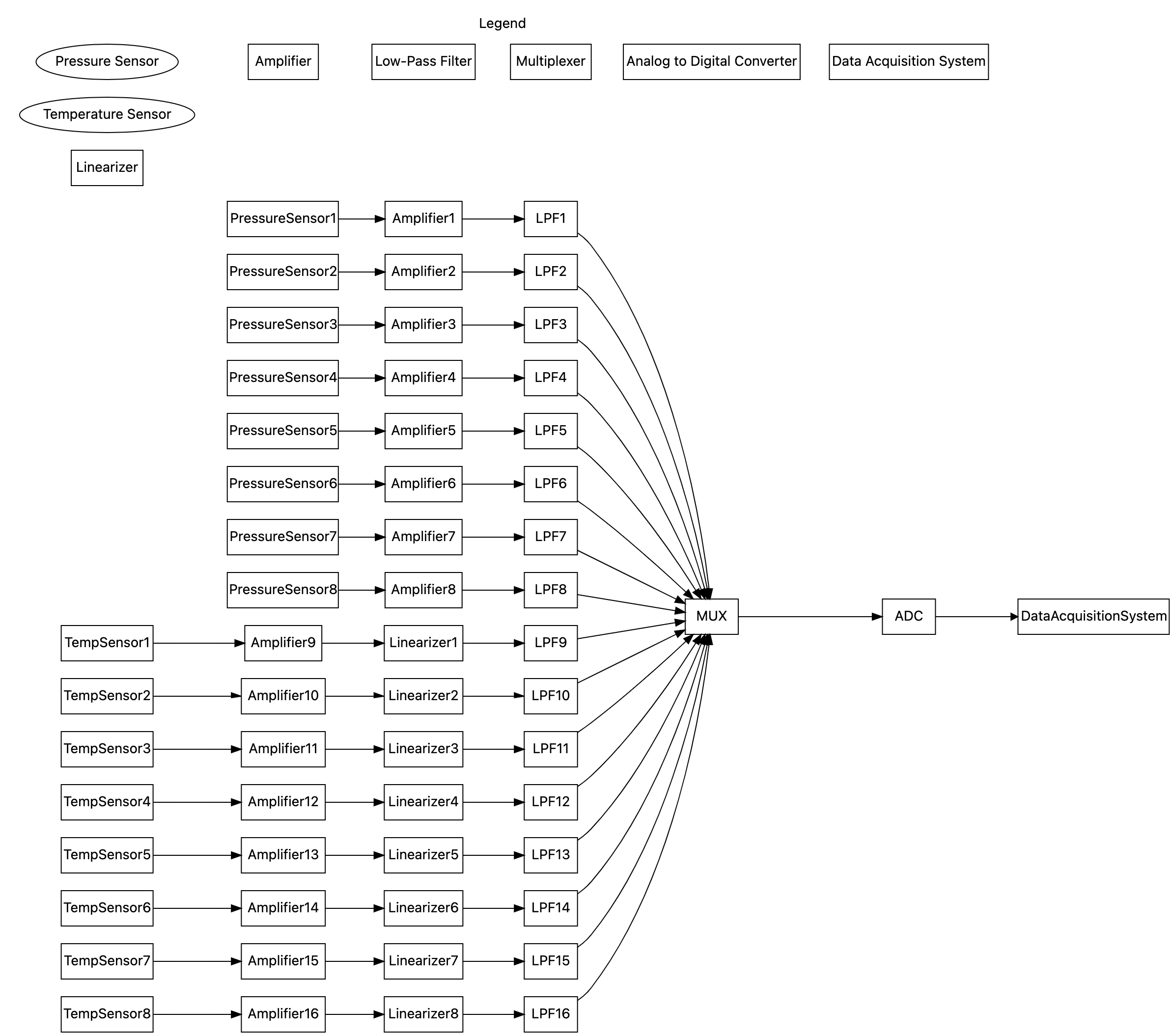}
    \label{fig:results_machine_massive}
\oautor
\end{figure}

The pressure and temperature monitor project caused some unique mistakes in the application, especially during the diagram generation. As there are 16 sensors, 8 of each, an option is to explicitly include the conditioning of each one. This might be considered the best approach, but given that the detailing phase is done for each block, this would cause the repetition of the detailing phase for blocks representing the same thing. This adds costs, processing time and allows the model to detail each clone block differently, which is undesirable. An example of this form of diagramming can be seen in Figure \ref{fig:results_machine_massive}. 

\begin{figure}[H]
    \caption{Diagram in which the application failed to successfully use arrays, found in iteration 16 with \textit{direct context}.}
    \centering
    \includegraphics[width=\columnwidth]{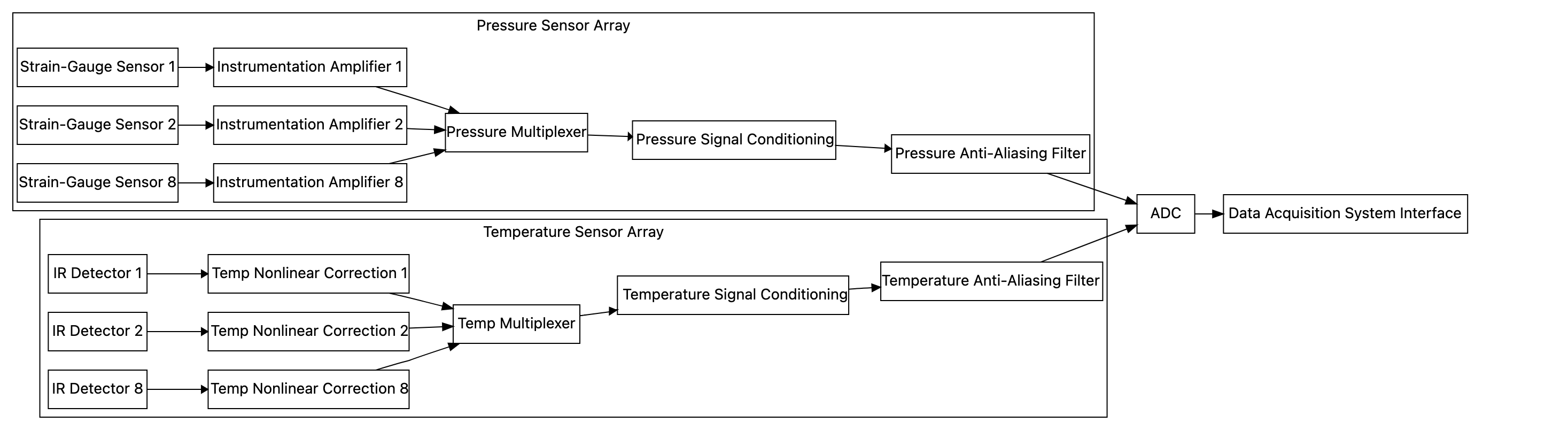}
    \label{fig:results_machine_bug}
\oautor
\end{figure}

This was identified during preliminary tests and therefore an instruction was included in the prompt to guide the model into preferring arrays in these cases, which ideally causes the solution to contain an architecture similar to Figure \ref{fig:best_machine}. However, in some cases, the model failed to represent the diagram using arrays, as can be seen in Figure \ref{fig:results_machine_bug}.

The most common theoretical mistake shown in this testbench, present in 37 out of 40 iterations, was omitting the use of a Wheatstone bridge for the strain gauges. Their sensitivity requires a differential signal and an instrumentation amplifier, in such a way that a solution containing only a non-inverting amplifier topology is not viable. Also, in 14 iterations an instrumentation amplifier was used for the single-ended temperature sensor. 

Given the existence of 16 channels, a multiplexer (MUX) was also needed, but in 12 iterations not all 16 channels were considered. Sometimes a 8-1 MUX was specified while in others a 2-1.

\section{Conclusion}
The goal of the paper was to develop a tool that provides  solutions to real project examples and evaluate its feasibility. Despite not all testbenches and iterations resulting in flawless solutions (rather the opposite), many of the limitations are due to GPTs capability - so far - to reliably execute such tasks.  

The tool proposes generating the entire synthesis at the system level, creating both the architecture and the implementation of each block. For this to happen, a series of interactions with GPT must be conducted, making the conversation extensive. It is known that language models struggle when the context of the message is too large, which may have worsened the design choices made by the model. 

The methodology used for the project was designed to develop a general-use tool (in the field of signal acquisition systems). The goal was not to test GPT's abilities specifically for the four selected testbenches, as overfitting scenarios were to be avoided. The prompts and conversational flow was not developed to be ideal for these four projects but for any project, regardless of whether the user's criteria were more quantitative or qualitative.

It is understood that the code implementation fulfilled its function well and managed to extract GPT's functionalities, even if it has limitations. However, the conversation history strategy caused consistency issues between blocks in multiple iterations. If it were possible to maintain the previous details in the conversation history without prohibitively increasing the context and losing information, some solutions would have likely achieved better results. Investigating how to mitigate this error could improve the tool.

When comparing the results between the \textit{direct context} and \textit{open context}, it's noticeable that the direct context provides more consistent solutions. However, that's to be expected given that it guarantees that all restrictions will be provided to the model, while the \textit{open context} depends on the model's ability to ask the appropriate questions at the right stage.  Therefore, the \textit{open context} should not be disregarded, but rather future research should work on improving the model's efficiency in asking questions.

Overall, the project was successful in structuring flows that enable LLMs (especially, GPT) to translate project requests in design decisions, even though its main conclusion is that much work is needed to turn LLM models into reliable tools to use in the data acquisition field.
Many of these limitations are still in regard to the model, but newer models are continually being released.

\addtolength{\textheight}{-12cm}   


\bibliographystyle{IEEEtran}  
\bibliography{ref.bib}

\end{document}

%% file: compare_contexts.tex
\begin{table}[H]
\centering
\caption{Comparison between the open and direct contexts.}
\label{tab:comparacao_contextos}

\begin{tabular}{ccc}
\hline
& \textbf{Direct} & \textbf{Open} \\ \hline

Sends the project description & \checkmark   &  \checkmark   \\

Sends project requirements in entry point &  \checkmark   &     \\
  
Can provide answers to questions &         & \checkmark  \\

Uses a GPT model     &    &   \checkmark  \\

\hline
\end{tabular}%

\oautor
\end{table}